\documentclass[12pt]{article}

\usepackage{amsmath}
\usepackage{amssymb}
\usepackage{amsthm}
\usepackage{graphicx}

\DeclareMathOperator*{\argmin}{arg\,min} 

\title{An Iterative Fingerprint Enhancement Algorithm Based on
Accurate Determination of Orientation Flow}

\author{Simant Dube\thanks{Supported by a research fellowship from Govt. of India}\\
Department of Computer Science and Engineering\\
Indian Institute of Technology\\
Kanpur 208016, India\\
Email: simantd (at) iitk.ac.in}

\begin{document}

\maketitle
\begin{abstract}
We describe an algorithm to enhance and binarize a fingerprint image. The algorithm is based on accurate determination of orientation flow of the ridges of the fingerprint image by computing variance of the neighborhood pixels around a pixel in different directions. We show that an iterative algorithm which captures the mutual interdependence of orientation flow computation, enhancement and binarization gives very good results on poor quality images.
\end{abstract}

\section{Introduction}
One of the most important tasks in an automatic fingerprint recognition system is enhancement of poor quality images~\cite{book:bhanu,nist_nbis}. Enhancement leads to better accuracy of the system as measured in terms of false rejection and false acceptance rates. Another important problem is binarization of the image into ridges and valleys, which allows for simple and fast binary image processing algorithms~\cite{nist_nbis}.

The starting point in most algorithms is that of accurate determination of orientation flow of a fingerprint image. In literature, one of the popular methods to compute the flow is based on gradient analysis, in which one computes second moment gradient matrix, also commonly known as structure tensor, in the neighborhood of each pixel to yield dominant orientation at each pixel~\cite{almansa_lindeberg,hong_jain,yu_xie_qi}. This approach works well if there is indeed a dominant orientation, but which is not the case in many regions of the image, especially if it is of poor quality.

Accurate determination of orientation flow in such problematic regions of fingerprint images is the first major contribution of this paper. Ours is a completely spatial approach which does not use gradients.

In literature, once the orientation flow is determined, authors have proposed several enhancement algorithms. Many use directional filtering and anisotropic diffusion methods~\cite{almansa_lindeberg,hong_jain,yu_xie_qi} or frequency domain methods~\cite{nist_nbis}. Recently, probabilistic techniques have been also proposed~\cite{lee_prabhakar}. 

We use directional filtering in our enhancement algorithm. Our filtering is faster as it uses short 1-D Gaussian filter as against 2-D Gaussian or Gabor filters commonly used in the past work. Furthermore, our algorithms for binarization and enhancement fit in the structure of our orientation flow algorithm so that they can be integrated.

Conventionally, the three tasks of orientation flow determination, enhancement and binarization tasks are performed sequentially. However it can be seen that these are mutually dependent on each other. For example, if we had an enhanced image, we could compute more accurate orientation flow, and vice versa. 

Motivated by this insight, we propose an iterative scheme and demonstrate that it gives excellent results. One important difference from the past work is that we perform binarization before enhancement. We demonstrate the performance of the proposed projection based iterative algorithm through examples.

\section{Second Moment Matrix Based Approach}
Let $I({\bf p})$ be gray level intensity of a fingerprint image at location ${\bf p} \in \mathbb{R}^2$. Define a neighborhood $\cal N$ around ${\bf p}$. Compute the second moment matrix of gradients of pixel intensity in this neighborhood
\begin{equation}
A = \sum_{{\bf q} \in {\cal N}} \bigtriangledown I({\bf q}) \bigtriangledown I({\bf q})^T.
\label{eq:second_moment}
\end{equation}
Let the eigen-values of $A$ be $\lambda_1 \geq \lambda_2 \geq 0$. Let the orientation angle of the dominant eigen-vector ${\bf v_1}$ be $\theta = \tan^{-1} ({\bf v_1}_{y}/{\bf v_1}_x)$. The angle $\theta$ is the orientation direction at pixel $p$, see~\cite{book:jahne}. Equivalently,
\[\theta = \frac{1}{2} \tan^{-1} \left( 
			\frac{\sum_{{\bf q} \in \cal N} 2 (\bigtriangledown I({\bf q}))_x (\bigtriangledown I({\bf q}))_y}
			{\sum_{{\bf q} \in \cal N} ({(\bigtriangledown I({\bf q}))_x}^2 - {(\bigtriangledown I({\bf q}))_y}^2)}\right)
			.\]

Often $\cal N$ is weighted by a Gaussian function so that pixels closer to ${\bf p}$ have
greater contribution.

\begin{figure}
\centering
\includegraphics{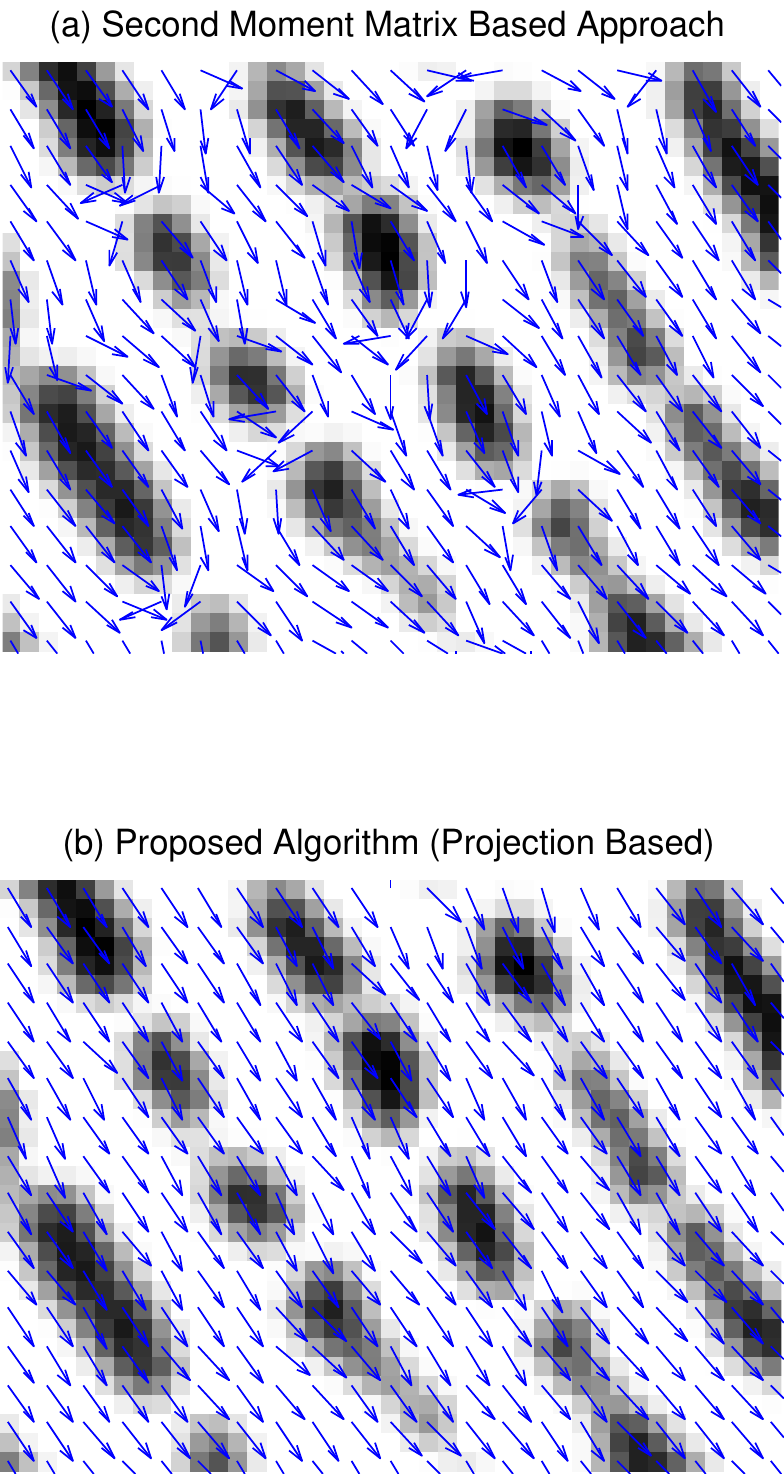}
\vspace{0.5in}
\caption{Orientation flow determination. (a) Well-known method based on second moment gradient matrix. (b) Our proposed algorithm on the same area of the image.}
\label{fig:compare}
\end{figure}

We show the problem with this popular approach in Figure~\ref{fig:compare}~(a) in which we show a $25 \times 30$ subregion of a fingerprint image captured by an optical sensor. At many pixels
there is no well defined dominant orientation due to overall poor quality of the image, which then leads to inaccurate orientation flow computation. We will show in next section how we can compute an accurate orientation flow as shown in Figure~\ref{fig:compare}~(b).

\section{Projection Based Iterative Algorithm}

\subsection*{Orientation Flow}
Informally, for any point in the image, we take projections in its neighborhood in different directions and find the direction which gives minimum variance in pixel values.

\begin{figure}
\centering
\includegraphics{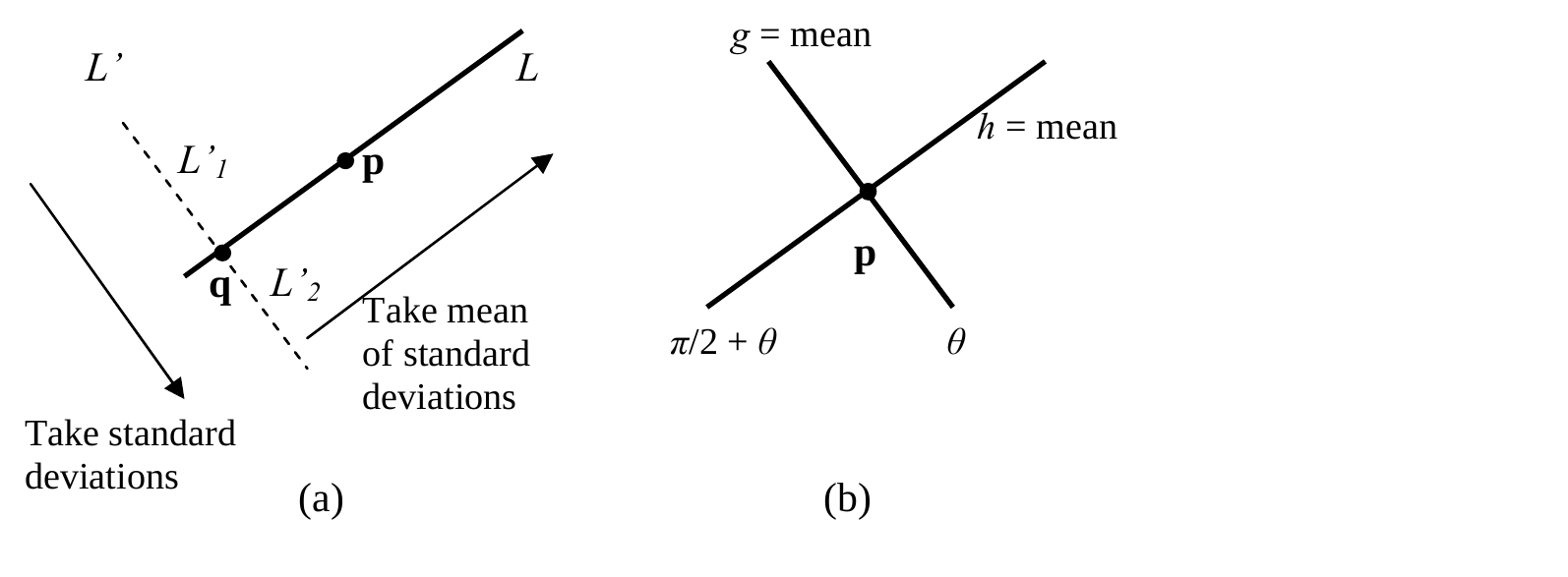}
\vspace{-0.5in}
\caption{(a) Computation of dominant orientation at pixel $\bf p$ in the fingerprint image. (b) Binarization of the pixel $\bf p$.}
\label{fig:projection}
\end{figure}

Consider neighborhood $\cal N$ around ${\bf p}$. Let $L({\bf p}, \alpha) \subset {\cal N}$ be the line segment passing through $\bf p$ at an angle $\alpha$. Let ${\bf q} \in L({\bf p}, \alpha)$. Line $L' = L({\bf q}, \alpha+\pi/2) \subset {\cal N}$ is the line segment passing through ${\bf q}$ which is perpendicular to $L({\bf p}, \alpha)$, see Figure~\ref{fig:projection}~(a).

Denote the standard deviation of pixel values on this perpendicular line as
\[\sigma({\bf q}) = \sigma(L') = \sqrt{{\rm Variance}\{I({\bf r}):{\bf r} \in L'\}}.\]
Take the mean of the standard deviations over all such lines
\[\mu_\alpha = {\rm mean}\{\sigma(L') : L'= L({\bf q}, \alpha+\pi/2),\; {\bf q} \in L({\bf p}, \alpha)\}.\]
Now we define {\it dominant orientation\/} at $\bf p$ to be the angle which gives minimum average of standard deviations over all angles $\alpha$, plus $\pi/2$:
\begin{equation}
\theta = \frac{\pi}{2} + \argmin_\alpha \{\mu_\alpha\}.
\label{eq:min_alpha}
\end{equation}

We found that the best results are obtained when we divide the line segment
$L'$ into two halves $L_1'$ and $L_2'$ which meet at ${\bf q}$, as shown in Figure~\ref{fig:projection}~(a), and compute standard deviations for these two line segments, and redefine $\sigma$ above as
\[\sigma({\bf q}) = \min\{\sigma(L_1'), \sigma(L_2'), \sigma(L')\}.\]
The reason why this gives better result is because even when a point is at the end of a ridge, we need to look at only the half of the neighborhood which is towards the ridge and ignore the other half.

See Figure~\ref{fig:compare}~(b) for results of our proposed algorithm.

We mention few speedup tricks we employed to make the algorithm fast:
\begin{enumerate}
\item
The orientation flow is not computed at all points ${\bf p}$, but only at subsampled set. We computed it at every other pixel in $x$ and $y$ directions yielding speedup factor of 4. At other positions, one can interpolate the angle.
\item
The minimum in Equation~\eqref{eq:min_alpha} is obtained by considering $\alpha$ from
0 to $\pi$ in increments of $\pi/8$. Then we refine this estimate $\theta$ by considering smaller increments of $\pi/32$ in the range $[\theta-\pi/16, \theta+\pi/16]$.
\item
We rotate the image once for each angle $\alpha$. Then we pre-compute $I({\bf x})^2$ for each pixel $x$. This makes computation of standard deviations fast.
\end{enumerate}

\subsection*{Binarization}
Informally, to compute the binary value of any pixel one has to check if the greyness values of pixels in the dominant direction is less or more than the values in the orthogonal direction. A related though not same idea is used in~\cite{nist_nbis}. For each point $\bf p$ with dominant orientation $\theta$, we compute the average greyness values along the line segments passing through it at angles $\theta$ and $\theta + \frac{\pi}{2}$, as shown in Figure~\ref{fig:projection}~(b):
\begin{eqnarray*}
g & = & {\rm mean}\{ I({\bf q}) : {\bf q} \in L({\bf p}, \theta)\},\\
h & = & {\rm mean}\{ I({\bf q}) : {\bf q} \in L({\bf p}, \theta + \frac{\pi}{2})\}.
\end{eqnarray*}
The binary value at $p$ is given as follows
\[ B({\bf p}) = \left\{ \begin{array}{ll}
											0, &  \; \mbox{if $g < h$} \\
											1, &  \; \mbox{otherwise}
											\end{array}
											\right. \]

\subsection*{Enhancement}
Given the orientation flow and the binary image, we can enhance the image as follows. Informally, for any point we apply a smoothing filter along the dominant direction at those neighboring points which have same binary values as the given point. Consider the sequence of points along line segment $L({\bf p}, \theta)$ which have binary value $B({\bf p})$. Apply a 1-D Gaussian filter on these points to get the enhanced image $J$ as
\[J({\bf p}) = {\cal G}(\sigma) * ({\bf q} \in L({\bf p}, \theta):B({\bf p}) = B({\bf q}))\]
where ${\cal G}(\sigma)$ is Gaussian filter of standard deviation $\sigma$, which we chose to be 3, and which is always normalized to have sum 1, since some ${\bf q}$ will not be considered if $B({\bf p}) \neq B({\bf q})$.

\subsection*{Nonlinear Isobrightness Contours}
As an advanced option of our algorithm, we generalize the line $L({\bf p},\theta)$ to a non-linear smooth curve $C({\bf p})$ which is the iso-brightness contour at $\bf p$. The points just next to $\bf p$ on this curve are:
\begin{eqnarray*}
{\bf p_1} & = & {\bf p} + (\cos \theta_{\bf p}, \sin \theta_{\bf p}),\\
{\bf p_2} & = & {\bf p} - (\cos \theta_{\bf p}, \sin \theta_{\bf p}).
\end{eqnarray*}
where we denote the dominant orientation at $\bf p$ by $\theta_{\bf p}$. The binarization and enhancement algorithm remain same except that we now use $C({\bf p})$. At most pixels, $L({\bf p},\theta)$ is very good approximation of $C({\bf p})$.

\subsection*{Iterative Algorithm}
\begin{figure}
\centering
\includegraphics{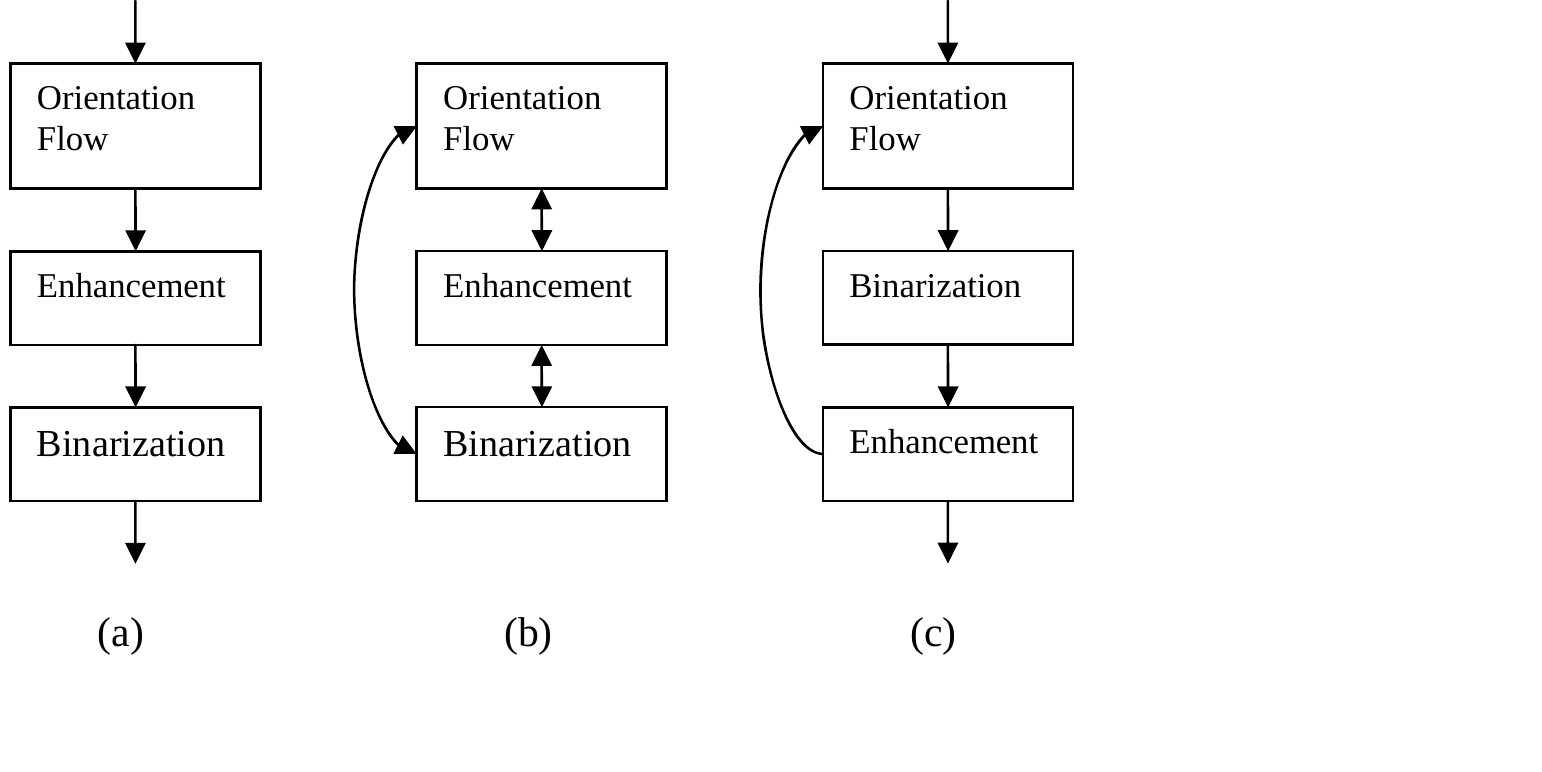}
\vspace{-0.7in}
\caption{(a) Conventional method to enhance and binarize a fingerprint image. (b) Mutual dependence of subtasks on each other. (c) Our implementation of mutual dependence using an iterative scheme.}
\label{fig:iterative}
\end{figure}

As we emphasized in the Introduction of this paper, we view orientation flow determination, binarization and enhancement to be three {\em mutually interdependent\/} components of overall algorithm, see Figure~\ref{fig:iterative}~(b). In Figure~\ref{fig:iterative}~(a), we show the conventional scheme. Therefore, we explore this novel approach by applying two iterations of our algorithm i.e. given an image $I$, once we have computed its orientation flow, binarized it, and enhanced it to give image $J$, then we repeat the whole process again with the enhanced image $J$ as the input, as illustrated in Figure~\ref{fig:iterative}~(c).

\section{Results}

\begin{figure}
\centering
\includegraphics[scale=1.05]{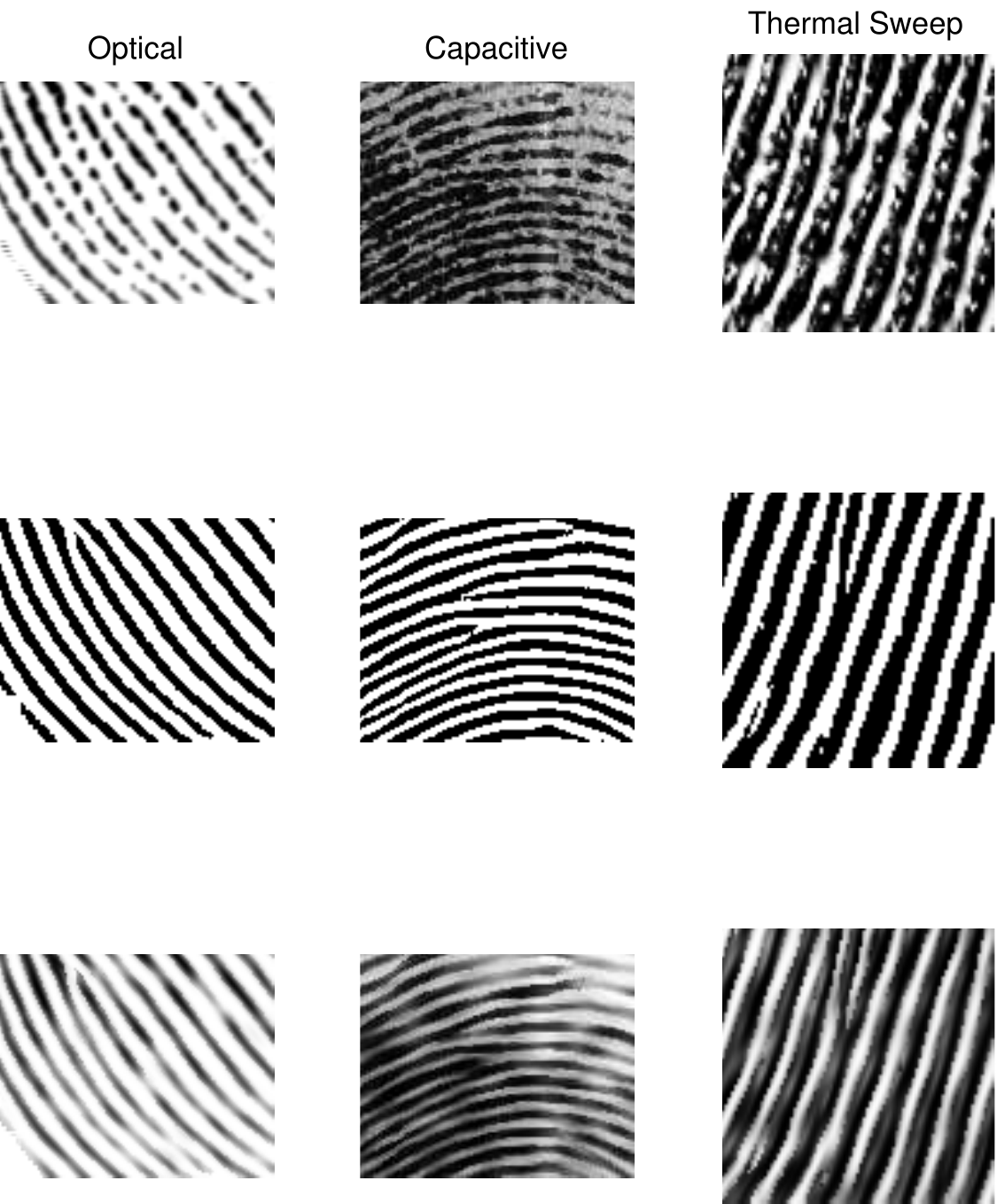}
\vspace{0.5in}
\caption{In the left column, we show results on an image captured by an optical sensor, and in the middle column, by a capacitive sensor, and in the right column, by a thermal sweep sensor. We show results in the selected subregions of the images. The top row shows the original images, the middle row shows the binarized images and the bottom row shows the enhanced images.}
\label{fig:results}
\end{figure}

In Figure~\ref{fig:results}, we show the results of our algorithm on three different kinds of fingerprint images, captured by optical, capacitive and thermal sweep sensors, see FVC 2002 and 2004 for the datasets~\cite{fvc}. The results on subregions of the images are shown in detail. We used linear approximation to iso-brightness contours for speed and because our experiments indicated that it gives results comparable to the non-linear case. Visual inspection of our results indicates that the algorithm is fairly robust as it works on all three types of sensors. We are able to reproduce the ridges very well, even in the case of capacitive sensor image which is of quite poor quality. One can detect minutiae in binarized images through the operation of morphological skeletonization or some other binary operations.

\section{Conclusions and Future Work}
We presented a novel, fast and robust algorithm to accurately compute the orientation flow of a fingerprint image and to binarize and enhance the image. Our algorithm for orientation flow significantly outperforms well-known method based on second moment matrix. We also showed very good results of binarization and enhancement algorithms on images captured by different kinds of sensors.

We believe ideas in this paper can be improved and optimized in future research in several ways. We are currently implementing a complete fingerprint recognition system based on this paper whose results will be described in a forthcoming publication. The mutual interdependence of different components of the algorithm as captured in the iterative algorithm needs to be understood more rigorously on a wide variety of fingerprint images.

\end{document}